# Segmentation et Interprétation de Nuages de Points pour la Modélisation d'Environnements Urbains

Jorge Hernández et and Beatriz Marcotegui
Mines ParisTech
CMM- Centre de morphologie mathématique
Mathématiques et Systèmes
35 Rue St Honoré 77305-Fontainebleau-Cedex, France
{hernandez,marcotegui}@cmm.ensmp.fr

**Résumé** Dans cet article, nous présentons une méthode pour la détection et la classification d'artefacts au niveau du sol, comme phase de filtrage préalable à la modélisation d'environnements urbains. La méthode de détection est réalisée sur l'image profondeur, une projection de nuage de points sur un plan image où la valeur du pixel correspond à la distance du point au plan. En faisant l'hypothèse que les artefacts sont situés au sol, ils sont détectés par une transformation de chapeau haut de forme par remplissage de trous sur l'image de profondeur. Les composantes connexes ainsi obtenues, sont ensuite caractérisées et une analyse des variables est utilisée pour la sélection des caractéristiques les plus discriminantes. Les composantes connexes sont donc classifiées en quatre catégories (lampadaires, piétons, voitures et «Reste») à l'aide d'un algorithme d'apprentissage supervisé. La méthode a été testée sur des nuages de points de la ville de Paris, en montrant de bons résultats de détection et de classification dans l'ensemble de données.

**Mots Clés :** Modélisation d'environnements urbains, Segmentation, Morphologie mathématique, Détection d'artefacts, Classification

***Abstract*** In this article, we present a method for detection and classification of artifacts at the street level, in order to filter cloud point, facilitating the urban modeling process. Our approach exploits 3D information by using range image, a projection of 3D points onto an image plane where the pixel intensity is a function of the measured distance between 3D points and the plane. By assuming that the artifacts are on the ground, they are detected using a Top-Hat of the hole filling algorithm of range images. Then, several features are extracted from the detected connected components and a stepwise forward variable/model selection by using the Wilk's Lambda criterion is performed. Afterward, CCs are classified in four categories (lampposts, pedestrians, cars and others) by using a supervised machine learning method. The proposed method was tested on cloud points of Paris, and have shown satisfactory results on the whole dataset.

## 1. Introduction

Depuis quelques années, grâce aux avancées technologiques dans le domaine de la télémétrie laser, plusieurs systèmes embarques sont apparus. Ces systèmes constituent un moyen rapide et précis de numériser en 3D (nuages de points) des environnements urbains. La segmentation et l'interprétation de cette information 3D est l'une des tâches les plus importantes dans la modélisation des environnements urbains.

La plupart des recherches sur l'exploitation de l'information 3D concernent l'approximation des façades par des plans. Dans [Dold06, Boulaassal07], des méthodes de croissance de régions sont utilisés pour extraire les surfaces planaires de façades et dans [Becker07, Stamos06] l'approximation de la façade est réalisée en utilisant l'algorithme de RANSAC. Madhavan et Hong, dans [Madhavan04], détectent la séparation de bâtiments et de routes. Goulette et al. [Goulette07] présentent une segmentation de nuages de points basée sur l'analyse de profils de points. Ils détectent la route, les façades et les arbres. Néanmoins, les deux dernières approches citées ont de fortes contraintes par rapport à leurs propres systèmes d'acquisition.

L'acquisition des données 3D est réalisée dans Paris, sans intervenir sur l'activité naturelle de la ville. Nous rencontrons différents types de véhicules (voitures, bus, motos, etc.), des piétons, ainsi que l'ensemble du mobilier urbain[1] (des lampadaires, des panneaux de signalisation, etc.) qui sont considérés comme des artefacts pour la modélisation de la scène. Ces artefacts devront être supprimés afin de faciliter la modélisation des façades et du sol.

Nos travaux se centrent sur la détection et l'interprétation des artefacts au niveau du sol comme : les voitures, les piétons, les lampadaires, etc. Cette détection a deux objectifs principaux: 1.- Le filtrage de ces structures pour faciliter l'étape de modélisation de bâtiments/façades et du sol et 2.- La réintroduction de certains éléments (lampadaires, signalisation), améliorant le réalisme visuel de la scène urbaine modélisée [Cornelis06].

Notre recherche s'inscrit dans le cadre du projet TerraNumérica du pôle de compétitivité mondial Cap Digital. Ce projet a pour objectif le développement d'une plate-forme de production et d'exploitation permettant la définition et la visualisation d'environnements urbains synthétiques. Cette plate-forme vise à augmenter la productivité et le réalisme de la modélisation.

L'article est organisé de la manière suivante. La section 2 présente les données sur lesquelles nous travaillons et la méthode de projection sur images. La section 3 illustre la segmentation par îlots pour travailler avec des nuages de points denses et gérer les limitations de mémoire. La section 3 décrit également une méthode qui segmente les données en façades et sol. Ensuite, la section 4 présente la détection d'artefacts basée sur l'image de profondeur. La section 5 décrit la méthode de classification des artefacts à partir d'une analyse des composantes connexes (*CC*s) issues de la section précédente. Des résultats expérimentaux sont illustrés dans la section 5. La dernière section expose les conclusions et les perspectives de ce travail.

---

[1] Sur la ville de Paris, il y a environ 200 types d'éléments de mobilier urbain. Information extraite du Cahier des Normes d'établissement et d'exploitation des plans de voirie de Paris.

## 2. Données

### 2.1. Systèmes d'acquisition et base de données

La méthode présentée a été testée sur des nuages de points acquis par deux systèmes mobiles différents (LARA 3D du CAOR – Mines ParisTech[2], et Stereopolis de l'IGN[3] correspondant à approximativement 2 kilomètres (une trentaine de pâtés de maisons) de rues du 5ème arrondissement de Paris. La Figure 1 présente une image panoramique de la zone de test.

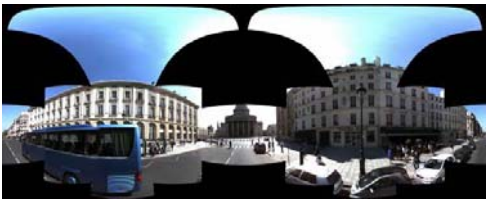

**Figure 1** : Vue panoramique de la rue Soufflot dans zone d'acquisition. Données © IGN.

Le système mobile LARA3D se compose de deux capteurs de perception, un télémètre laser et un appareil photo équipé d'un grand-angle (fisheye) et équipé de capteurs de géo-localisation (GPS, IMU, odomètres) [Brun2007]. Le scanner couvre une zone de 270 degrés, en acquérant les deux côtés de la rue, la toiture de la façade et le sol (trottoir et route).

Le système mobile Stereopolis se compose de 16 appareils photos de haute résolution, de deux télémètres laser et des dispositifs de géo référencement. Chaque télémètre laser a 80 degrés de champ de vision. Les télémètres sont dirigés vers la façade avec des angles d'incidence de 90 et à 45 degrés.

### 2.2. Image de Profondeur et image d'accumulation

Les méthodes que nous avons développées s'appuient sur une image de profondeur et une image d'accumulation des nuages de points. Les images sont générées en utilisant une caméra virtuelle (***P***) positionnée sur le plan XY. L'image de profondeur [Foley95] enregistre la distance maximale des points 3D projetés sur le plan de la caméra et l'image d'accumulation compte le nombre de points (3D) qui sont projetés sur le même pixel de l'image de caméra.

La Figure 2 illustre un exemple de nuage de points et les images (profondeur et accumulation) sur lesquelles nous avons travaillé. Notez que dans les images, les valeurs maximales se trouvent dans la région des façades. Ce comportement corrobore le fait que les façades sont les structures les plus hautes et où la plupart des points sont projetés.

[2] caor.ensmp.fr/french/recherche/rvra/3Dscanner.php
[3] recherche.ign.fr/labos/matis/accueilMATIS.php

Afin d'éviter les problèmes d'échantillonnage, les valeurs de nombres de pixels par unité de longueur doivent être soigneusement choisies. Si elles sont trop petites, plusieurs points seront projetés sur les mêmes coordonnées image et il y aura une importante perte d'information. Par contre, si elles sont trop larges, la connexité de pixels, requise par nos méthodes, n'est pas assurée. Par conséquent, le choix idéal des dimensions est de *1:1*, autrement dit, un pixel pour chaque point 3D dans le plan de la caméra. Dans notre cas, les nuages de points ont une résolution approximative de *5 [cm]*, c'est la raison pour laquelle la résolution choisie est égale à *20 [pix/m]*.

## 3. Prétraitement

### 3.1. Segmentation en îlots

Pour faciliter le traitement des nuages de points denses, nous nous intéressons à la séparation des pâtés de maisons. Pour le découpage en pâtés de maisons, on peut exploiter l'information des images avec une résolution réduite de *4:1* (*5 [pix/cm]*) de manière à éviter les problèmes de mémoire. Pour cela, nous supposons que les différentes façades d'une même rue sont alignées. Cette hypothèse est vérifiée par les données dont nous disposons. L'algorithme utilise la transformée de Hough pondérée par la valeur d'intensité du pixel. Cette procédure nous permet de détecter la ligne de direction des façades sur l'image de profondeur (voir Figure 3(a)). Sur cette ligne nous analysons le profil des hauteurs des bâtiments. Nous détectons les variations de hauteur importantes, qui correspondent aux rues séparant les différents pâtés de maisons. Dans cette analyse, le profil est d'abord filtré pour réduire le bruit et faciliter la détection. La Figure 3(b) montre le profil original et le profil filtré. Ensuite, le profil est segmenté en utilisant la ligne de partage des eaux. Ces divisions sont tracées dans la direction perpendiculaire à la direction de la façade pour découper l'ensemble en tronçons de rue (Figure 3(c)). La segmentation est rétro-projeté pour découper le nuage de points 3D original (Figure 3(d)). Finalement, nous travaillons sur les sections découpées par pâtés de maison.

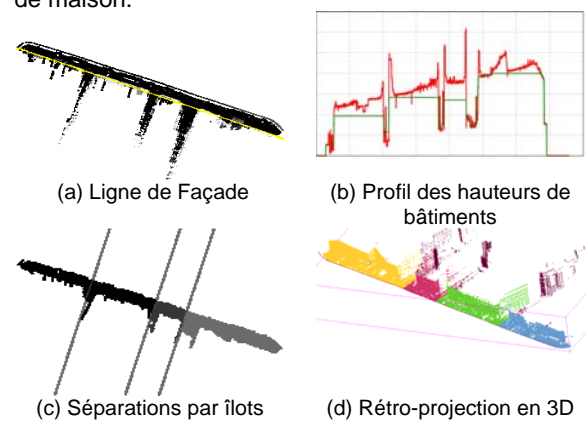

(a) Ligne de Façade    (b) Profil des hauteurs de bâtiments

(c) Séparations par îlots    (d) Rétro-projection en 3D

**Figure 3** : Exemple de la procédure de séparation par îlots.

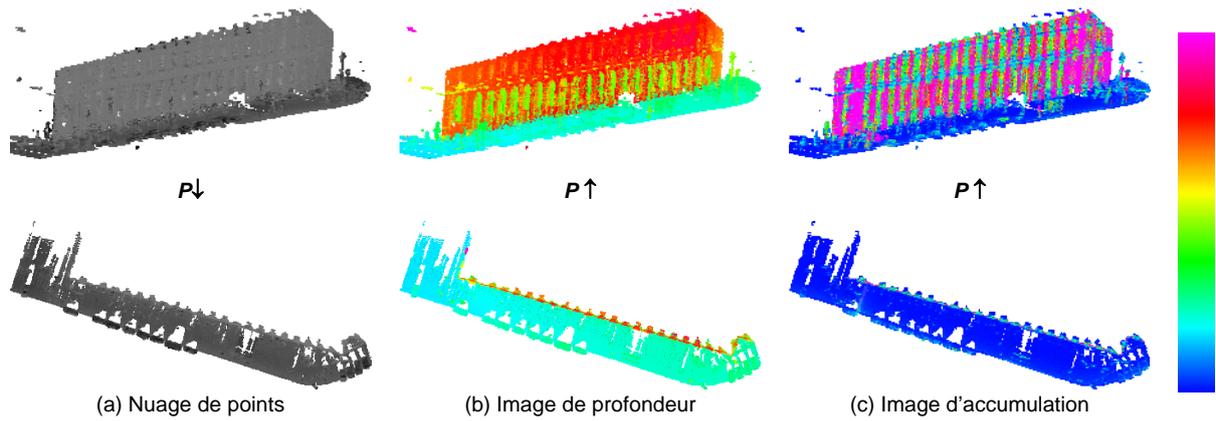

(a) Nuage de points  (b) Image de profondeur  (c) Image d'accumulation

**Figure 2** : Exemple des images de profondeur et d'accumulation. **(a)** Nuages de points - vue 3D et vue supérieure. Projection sur images (code couleurs au rang): **(b)** la valeur maximale – rang (*0,23651[mm]*), et **(c)** l'accumulation – rang (*0,121*[fois]). Données © IGN Tronçon de la Rue Soufflot côté pair.

### 3.2. Segmentation façades-sol

Une autre étape aussi intéressante à analyser est la segmentation de points 3D entre la façade et le sol (rue et trottoir). Cette segmentation est importante parce qu'elle facilite plusieurs étapes de la modélisation urbaine :

- La modélisation de la façade.
- L'extraction de la façade pour chaque bâtiment à partir d'une transformation de caméra.
- La détection d'artefacts au niveau du sol (voir section 4)

Nous travaillons avec les données déjà découpées en pâtés de maison et avec une résolution idéale d'image 20[pix/cm]. Néanmoins les méthodes présentées pourront être appliquées aux données brutes sous la contrainte de la disponibilité de mémoire.

Cette approche est basée sur une segmentation grossière de l'image de profondeur, réalisée au moyen d'un algorithme de zones quasi-plates (λ- zones plates) présenté par Meyer dans [Meyer98]. Deux pixels voisins *p* et *q* appartiennent à la même zone quasi plate, si leur différence de profondeur *f* est inférieure à une valeur λ :

$$\forall (p,q) \text{ voisins} : \left| f_p - f_q \right| \leq \lambda \qquad (1)$$

Etant donné que les variations de profondeur au niveau du sol sont faibles, la segmentation de l'image de profondeur avec une distance λ ≈ 1[m] réunira les pixels du sol dans une seule région en incluant les artefacts qui s'y trouvent.

Cependant, à cause des occlusions et des caractéristiques de systèmes d'acquisition, il y a des informations manquantes au niveau du sol. Ces données manquantes limitent la performance de la méthode parce que la connexité des pixels du sol n'est pas assurée. De manière générale, la taille des informations manquantes dépend principalement de la hauteur de l'obstacle, de la distance du capteur au sol et de la distance du capteur à l'obstacle, comme la Figure 4 l'illustre. L'image présente le processus d'acquisition par des systèmes mobiles d'un trottoir en pente et avec deux obstacles. Dans cet exemple, le sol est divisé en une région de rue par la voiture et deux régions de trottoir par le piéton. La Figure illustre également un profil obtenu par la projection du point 3D le plus haut (profil de profondeur de la vue supérieure).

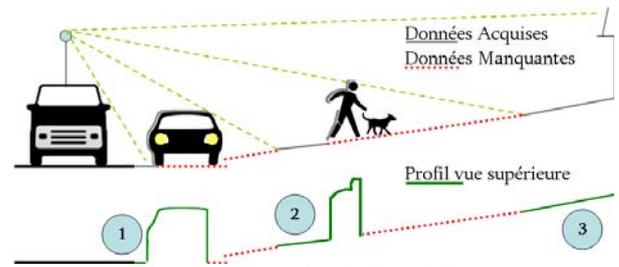

**Figure 4** : Illustration de l'acquisition.

En morphologie, il existe une technique qui permet le bouchage des régions dénommées trous. Un trou est un ensemble de pixels dont le minimum n'est pas connecté au bord de l'image *I* [Soille03]. L'algorithme de remplissage de trous consiste donc à supprimer les minima qui ne sont pas liés au bord de l'image. Il utilise une reconstruction morphologique par érosion où le marqueur est une image à la valeur maximale partout sauf sur son bord (voir équation (2)).

$$g = \text{FILL}(f) = R_f^e(f_m) \quad \text{où} \qquad (2)$$

$$f_m = \begin{cases} f_p & \text{si p est au bord} \\ \max(f) & \text{autrement} \end{cases}$$

La Figure 5 illustre un exemple de remplissage de trous *g* d'un profil *f*.

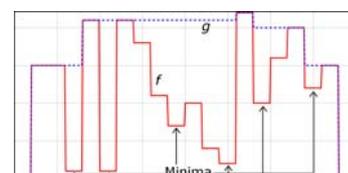

**Figure 5** : Remplissage de trous 1D.

Ainsi, en utilisant l'algorithme de remplissage, on peut donc boucher les trous qui ne touchent pas le bord. Malgré cela, comme on peut apprécier dans la Figure 6(b), seulement quelques trous ont été remplis. Dans le cas des données de systèmes mobiles, pour une scène avec beaucoup d'obstacles (par exemple, des voitures garées) ou de grands obstacles (des lampadaires), les trous sont connectés au bord de l'image.

Pour résoudre ce problème, nous proposons de relier les régions du sol entre elles, en utilisant le chemin de la distance minimale qui les séparent. Autrement dit une région sera reliée à sa voisine la plus proche. Le résultat final des connexions est illustré dans la Figure 6(c). Maintenant, les régions et ses connexions forment de nouveaux trous qui doivent être bouchés. Le remplissage s'applique bassin par bassin de manière indépendante afin de réaliser une estimation locale de la région à combler. La Figure 6(d) montre le nouveau remplissage de l'image de profondeur.

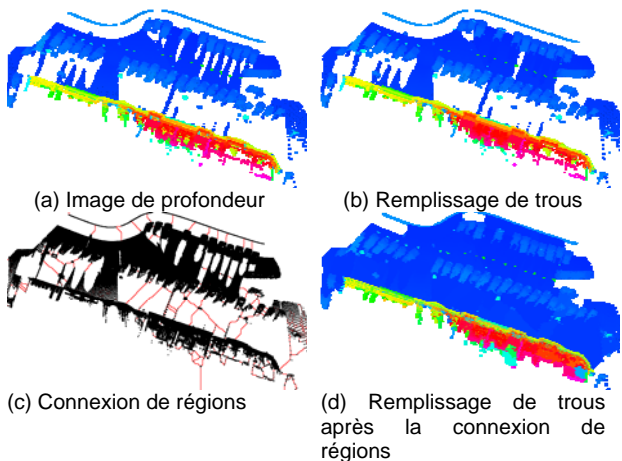

(a) Image de profondeur  (b) Remplissage de trous

(c) Connexion de régions  (d) Remplissage de trous après la connexion de régions

**Figure 6** : Estimation des pixels manquants en utilisant le remplissage de trous.

En utilisant l'image comblée, nous pouvons appliquer la segmentation de zones quasi-plates, en obtenant comme résultat les images de la Figure 7(a). En choisissant la région la plus grande nous obtenons le masque du sol. Cette première région présente quelques trous produits par des éléments très hauts comme les lampadaires. Afin de récupérer ces éléments, on assigne au masque du sol tous ceux qui sont contenus dans le premier masque. Le masque final est rétro-projeté sur le nuage de points, permettant la segmentation de nuages de points 3D de façades et points du sol (voir Figure 7(b)).

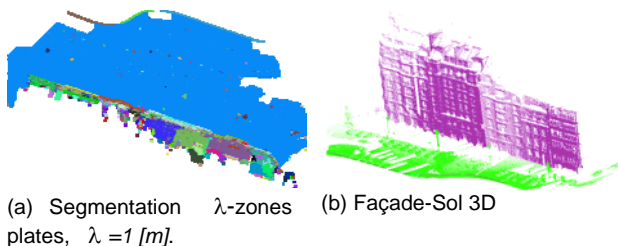

(a) Segmentation λ-zones plates, λ =1 [m].  (b) Façade-Sol 3D

**Figure 7** : Segmentation Façade-Sol.

### 4. Détection d'artefacts

Nous faisons l'hypothèse que les artefacts se trouvent sur le sol. C'est pourquoi, nous utilisons la séparation façade - sol pour extraire un masque du sol. Ensuite, nous ne traitons que les artefacts à l'intérieur de ce masque. Comme image de départ nous utilisons l'image de profondeur comblée que nous avons utilisée pour la détection du sol (voir la Figure 6(a)). Avec cette étape, l'image de profondeur est plus homogène car cette transformation réduit également le bruit sur les artefacts. La méthode de détection est basée sur le chapeau haut de forme de l'algorithme de remplissage de trous (voir équation 3).

$$\text{FTH}(f) = g - f = \text{FILL}(f) - f \qquad (3)$$

Analysons l'illustration de la Figure 4. Les deux artefacts sont des bosses sur le profil. Ainsi, si nous inversons le profil, ces bosses deviennent des trous. En utilisant l'opérateur de chapeau de forme, après un remplissage de trous du profil inversé, nous obtenons une détection d'artefacts avec une estimation de leur hauteur. Comme on peut observer, dans l'illustration de la Figure 8, les deux artefacts sont correctement détectés.

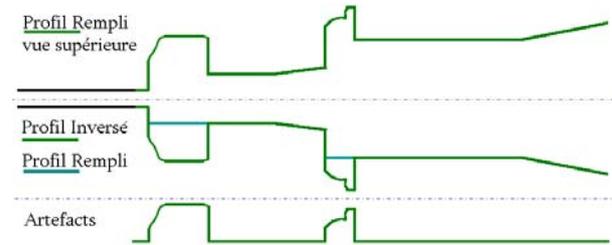

**Figure 8** : Détection d'artefacts de l'illustration.

Même si on a décrit la méthode sur un profil, cette technique est utilisée en 2D sur l'image de profondeur remplie. Néanmoins deux détails additionnels doivent être considérés :

1. Une fois que l'image est inversée, le fond de l'image de profondeur (noir) devient la valeur maximale. De cette manière, toute l'information de profondeur est un grand trou. Pour résoudre ce problème nous inversons seulement les points différents de zéro (les pixels à zéro restent à zéro).

2. Les artefacts qui se trouvent à la frontière, spécialement les voitures, manquent de l'information latérale à cause de la proximité du capteur des objets. En faisant le chapeau haut de forme, l'artefact n'aura que l'information de la partie haute. Le problème est corrigé en ajoutant, autour du masque du sol, un bord à la hauteur minimale de l'image de profondeur.

Finalement, un seuil de *10 [cm]* est appliqué à l'image résultat pour éliminer les artefacts produits par la rugosité du sol et les surfaces bruitées. La méthode est résumée par le diagramme de la Figure 9. La méthode présente deux sorties : l'image d'artefacts et une image que nous appellerons le marqueur du sol. Comme on peut observer, ce marqueur est le même masque initial sans les pixels qui appartiennent aux artefacts. Les images de sortie peuvent être rétro-projetées aux nuages de points pour réaliser la segmentation 3D comme la Figure 10 l'illustre.

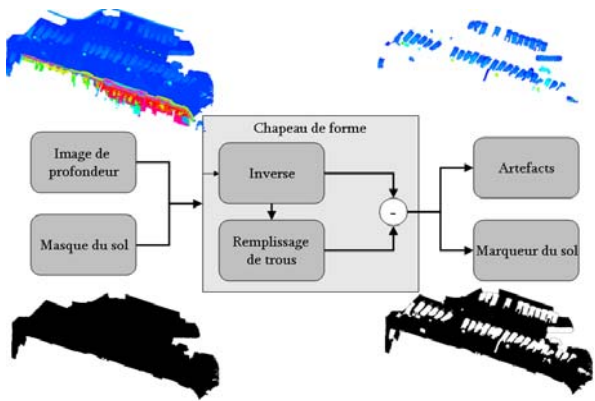

**Figure 9** : Les étapes de la détection d'artefacts.

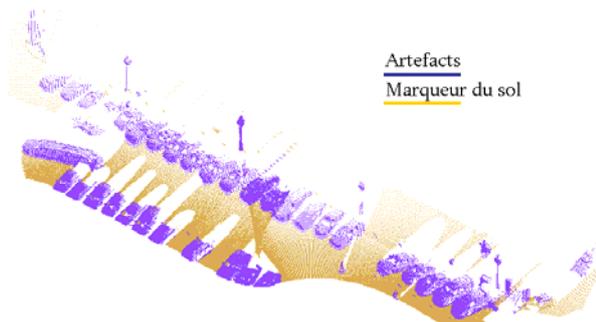

**Figure 10** : Détection des artefacts 3D.

## 5. Classification d'artefacts

### 5.1. Artefacts connectés

Les artefacts sont bien détectés, ce qui permet de filtrer correctement les données pour une modélisation des façades et du sol. Par contre, leur classification nécessite que chaque artefact soit une composante connexe (*CC*) séparée des autres. Néanmoins, en utilisant une simple labellisation, plusieurs artefacts peuvent rester connectés entre eux formant une seule *CC*. La Figure 11 illustre plusieurs situations d'un *CC* avec plusieurs artefacts, des piétons en train de toucher un lampadaire (Figure 11(a)), deux voitures trop proches (Figure 11(b)) et un piéton en train d'ouvrir la porte de sa voiture (Figure 11(c)).

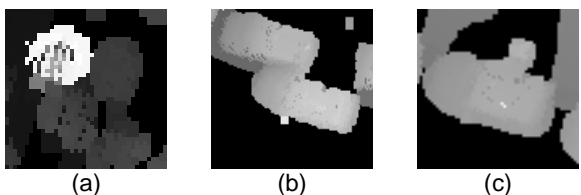

(a) (b) (c)
**Figure 11** : Des artefacts connexes.

Nous avons décrit un artefact comme une bosse, ainsi le nombre d'artefacts connectés dans une même *CC* dépend du nombre de bosses qu'elle contient. Ces bosses peuvent être exprimées en termes de nombres de maxima de la *CC*. Cependant, une *CC* peut avoir plusieurs maxima causés par de petits pics et de la rugosité dans la partie supérieure de l'artefact. Pour extraire les maxima les plus significatifs, deux filtres morphologiques sont utilisés. Une ouverture par surface [Vincent94] et un opérateur *h*-Maxima [Soille03].

L'opérateur *h*-Maxima élimine les maxima qui ont une profondeur inférieure ou égale à un seuil *h*. Il utilise une reconstruction morphologique par dilatation où le marqueur est une image à laquelle on a soustrait une constante *h* de *10[cm*].

Une fois les artefacts filtrés, on utilise la ligne de partage des eaux de l'image de gradient avec des marqueurs. Les marqueurs sont les maxima et le fond de l'image filtrée. La Figure 12 illustre trois exemples de la séparation d'artefacts connectés, en montrant la performance de la méthode.

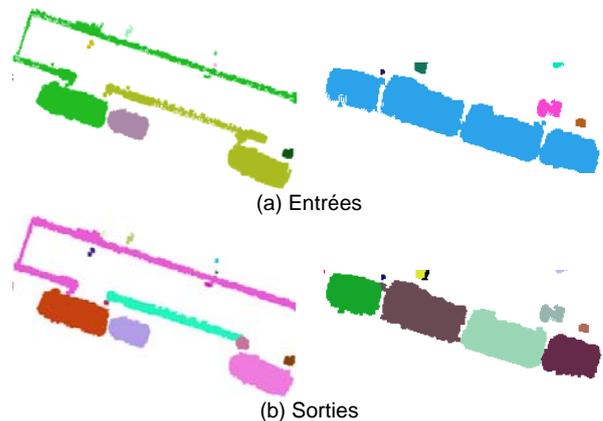

(a) Entrées

(b) Sorties
**Figure 12** : (a) Détection initiale et (b) Séparation d'artefacts.

### 5.2. Sélection de caractéristiques

Chaque *CC* de l'image est analysée afin de déterminer le type d'artefact. Quatre catégories d'artefacts sont analysées des lampadaires, des voitures, des piétons et le reste (des poubelles, des motos, des parc-mètres, des abris bus, etc.).

Les mesures calculées sont :

1. La moyenne, l'écart-type, le maximum, le minimum et le mode de la hauteur [mètres] ;
2. La moyenne, l'écart-type, le maximum, le minimum et le mode de l'accumulation [fois] ;
3. La surface de la composante connexe. Initialement, cette caractéristique est calculée en pixels. Toutefois, en utilisant l'information de la caméra, la valeur est transformée en mètres carrés. Cette transformation homogénéise les éléments qui appartiennent aux mêmes classes dans l'ensemble d'images (les images de profondeur sont générées de manière indépendante).

Nous avons annoté la base de données pour établir une vérité terrain de 442 artefacts : (67 voitures, 33 lampadaires, 198 piétons et 144 du reste). Pour réduire notre espace de onze caractéristiques, nous utilisons une étape de sélection de variables. L'idée est de sélectionner les variables au fur et à mesure. Les variables sont sélectionnées en utilisant le critère de lambda de Wilk [Mardia79,Roever09]. La Figure 13 montre les variables ordonnées selon cette méthode.

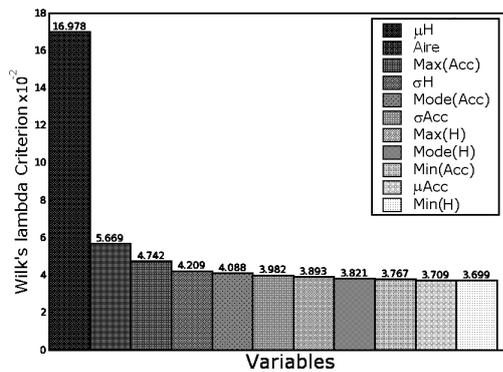

**Figure 13** : Critère de lambda de Wilk de l'étape de sélection de variables (caractéristiques ordonnées).

Une distribution de classes avec les trois premières caractéristiques est illustrée par la Figure 14; où la surface est la variable caractéristique des voitures et la moyenne de la hauteur celle des lampadaires. Néanmoins, la classe « Reste » est entrecroisée principalement avec la classe des piétons. Si nous fixons à *0,01* la valeur maximale d'erreur autorisée (*p-value*) par le modèle réduit par rapport au modèle avec toutes les variables, nous pouvons réduire à seulement six caractéristiques les plus importantes: la moyenne de la hauteur, la surface, le maximum de l'accumulation, l'écart-type de la hauteur, le mode de l'accumulation et l'écart-type de l'accumulation.
.

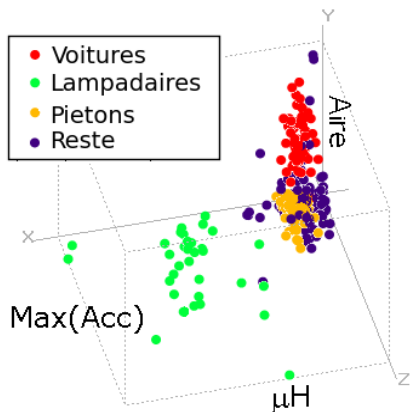

**Figure 14** : Distribution des classes dans l'espace la moyenne de la hauteur vs la surface vs le maximum de l'accumulation.

### 5.3. Classification

Avec les caractéristiques sélectionnées, nous éliminons d'abord les *CC*s de petite taille (*10 [pix]*) et qui ont moins de 3 points d'accumulation. Puis nous utilisons une méthode de classification supervisée, nommée séparateurs à vastes marges (*support vector machine* SVM) [Chang01,Dimitriadou09]. La classification a été validée en utilisant la validation croisée (*K-fold cross validation, K= 10*).

Le résultat de la classification est représenté dans la matrice de confusion de la Table 1. La table montre de bons résultats de *classification* avec l'utilisation de seulement six caractéristiques. Tous les lampadaires ont été bien classifiés, et les voitures et les piétons ont de bons taux de classification, de 91,04% et 96,97%. Néanmoins, une grande quantité des artefacts de la classe « Reste » ont été classifiés comme des piétons, un exemple de ce problème est présenté avec les horodateurs qui sont des structures avec des caractéristiques équivalentes aux piétons (hauteur et surface similaires). L'erreur totale de classification est de 14,93%.

|  | Voitures | Lampadaires | Piétons | Reste |
|---|---|---|---|---|
| Voitures | 91,04 | 0,0 | 0,0 | 8,96 |
| Lampadaires | 0,0 | 100,0 | 0,0 | 0,0 |
| Piétons | 0,0 | 0,0 | 96,97 | 3,03 |
| Reste | 7,64 | 1,39 | 28,47 | 62,50 |

**Table 1** : Matrice de confusion de la classification d'artefacts

Les Figure 15 et 16 montrent les résultats de la méthode présentée dans cet article. La méthode de détection présente de bons résultats dans l'ensemble de données testées.

### 7. Conclusions et Travaux futurs

Une méthode de segmentation et d'interprétation de nuages de points au niveau du sol d'environnements urbains a été présentée. La détection d'artefacts (même sans classification) est utile pour filtrer les données et faciliter la modélisation des façades et du sol. La méthode permet la détection de lampadaires, de voitures et de piétons ainsi que la localisation d'autres artefacts non encore reconnus. Nous travaillons actuellement sur la détection de panneaux de signalisation et nous prévoyons d'étendre la méthode à la classification d'autres artefacts tels que les motos et les poubelles. L'introduction d'autres classes réduira le taux de fausses classifications de la classe « Reste ». Des tests à plus grande échelle sont prévus dans le cadre du projet TerraNumerica.

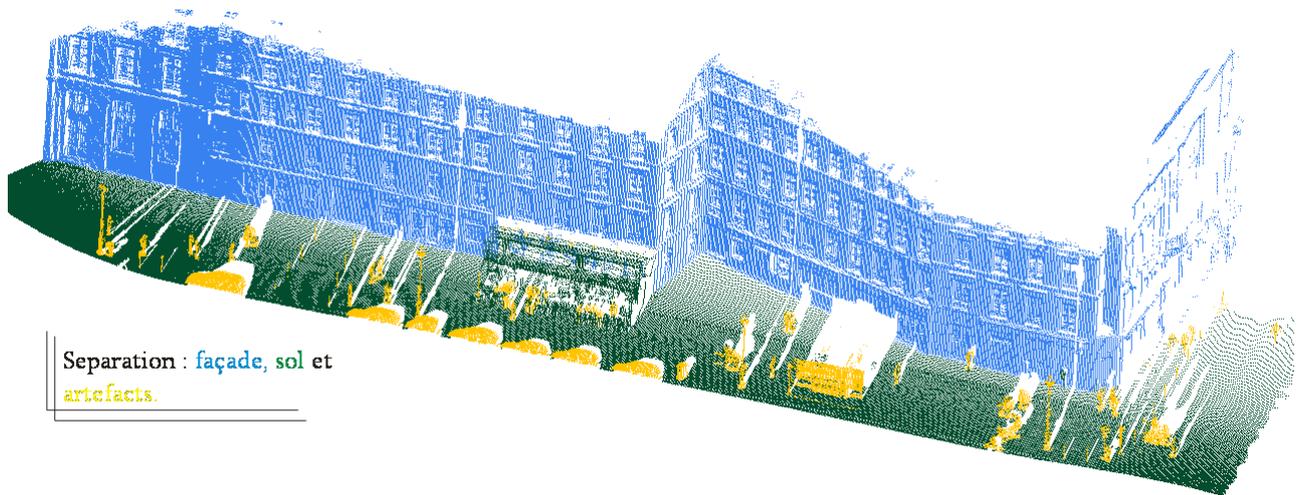

**Figure 15** : Détection d'artefacts et filtrage de données façade et sol.

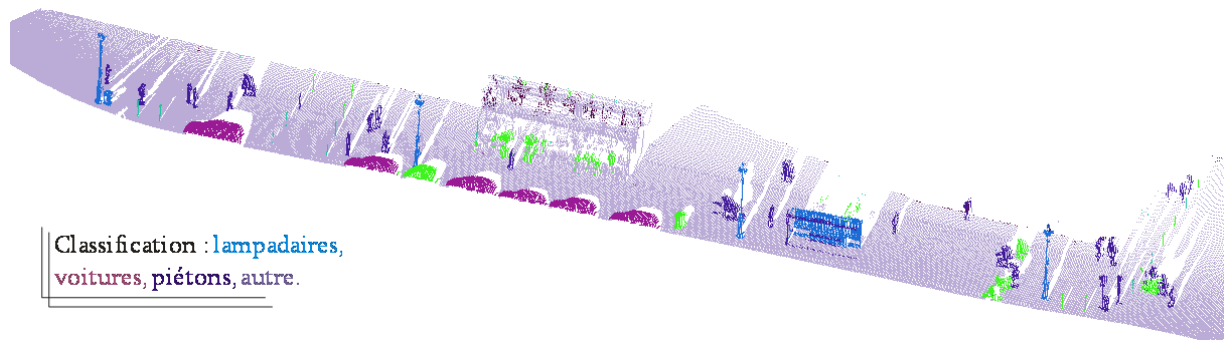

**Figure 16** : Classification d'artefacts en quatre classes.